\pdfoutput=1

\documentclass[11pt]{article}

\usepackage[final]{acl}

\usepackage{times}
\usepackage{latexsym}

\usepackage[T1]{fontenc}

\usepackage[utf8]{inputenc}

\usepackage{microtype}

\usepackage{inconsolata}

\usepackage{multirow}
\usepackage{amsmath}
\usepackage{amssymb}
\usepackage{amsfonts}
\usepackage{multirow}
\usepackage{multicol}
\usepackage{array}
\usepackage{bm}
\usepackage{booktabs}
\usepackage{amsmath}
\usepackage{graphicx}
\usepackage{algorithm} 
\usepackage{algorithmic}
\usepackage{subfigure}

\title{Leveraging Estimated Transferability Over Human Intuition for\\ Model Selection in Text Ranking}

\author{Jun Bai$^{1}$\quad\quad Zhuofan Chen$^{1}$\quad\quad Zhenzi Li$^{1}$\quad\quad Hanhua Hong$^{2}$\\ {\bf Jianfei Zhang$^{1}$}\quad\, {\bf Chen Li$^{1}$}\quad\quad {\bf Chenghua Lin$^{2}$}\quad\, {\bf Wenge Rong}$^{1}$\\
    $^{1}$ School of Computer Science and Engineering, Beihang University, China\\
    $^{2}$ Department of Computer Science, University of Manchester, United Kingdom\\   \{ba1\_jun, zhuofanchen, zhenzil, zhangjf, chen.li, w.rong\}@buaa.edu.cn\\
    hanhua.hong@postgrad.manchester.ac.uk, chenghua.lin@manchester.ac.uk
}

\begin{document}
\maketitle
\begin{abstract}
Text ranking has witnessed significant advancements, attributed to the utilization of dual-encoder enhanced by Pre-trained Language Models (PLMs). Given the proliferation of available PLMs, selecting the most effective one for a given dataset has become a non-trivial challenge. As a promising alternative to human intuition and brute-force fine-tuning, Transferability Estimation (TE) has emerged as an effective approach to model selection. However, current TE methods are primarily designed for classification tasks, and their estimated transferability may not align well with the objectives of text ranking. To address this challenge, we propose to compute the expected rank as transferability, explicitly reflecting the model's ranking capability. Furthermore, to mitigate anisotropy and incorporate training dynamics, we adaptively scale isotropic sentence embeddings to yield an accurate expected rank score. Our resulting method, Adaptive Ranking Transferability (AiRTran), can effectively capture subtle differences between models. 
On challenging model selection scenarios across various text ranking datasets, it demonstrates significant improvements over previous classification-oriented TE methods, human intuition, and ChatGPT with minor time consumption.
\end{abstract}

\section{Introduction}
\label{sec: intro}
Ranking relevant documents based on user queries has gained prominence as an essential component for Information Retrieval (IR) \citep{DBLP:conf/aaai/00010WWL24/ir}, Retrieval-Augmented Generation (RAG) \citep{DBLP:journals/corr/abs-2401-14887/rag}, etc.
As the volume of online content continues to grow, developing efficient and effective text ranking techniques has become increasingly imperative. 
Fortunately, the advancements in Pre-trained Language Models (PLMs) have spurred the development of dual-encoder capable of meeting these demands \citep{zhao2024dense}.
Usually, practitioners select the PLM to be employed by their intuition.
However, such decision-making becomes unreliable as the number of PLMs rapidly grows \citep{DBLP:journals/corr/abs-2402-02314/human_poor_MS}, introducing a new challenge: the accurate and automatic selection of the most potent PLM to achieve the desired performance on given datasets, namely Model Selection (MS) \citep{DBLP:conf/emnlp/BaiZLHXLR23/emnlp_survey}.

Early approaches to MS adopted a brute-force method, fine-tuning each model to identify the optimal one \citep{DBLP:conf/icpr/ChoiKJG20}, which suffers from huge computing costs.
Hence, it is preferable to estimate model fine-tuning result with computational efficiency. 
In pursuit of this goal, Transferability Estimation (TE) has garnered increasing attention in recent years, merely leveraging model-encoded features and labels \citep{DBLP:conf/eccv/AgostinelliPUMF22}.
Currently, advances in TE have primarily been developed for classification tasks \citep{DBLP:conf/emnlp/BassignanaMZP22/logme_nlp}. 
These methods approximate the log-likelihood from features to labels, aligning with the classification objectives well and demonstrating promising results on classification datasets \citep{DBLP:conf/emnlp/BaiZLHXLR23/emnlp_survey}.
However, the text ranking task fundamentally differs from classification task in which the matching scores for relevant query-document pairs are expected to be higher than those for irrelevant pairs \citep{DBLP:journals/tacl/LuanETC21}.
As a result, applying current classification-oriented TE methods directly to text ranking is sub-optimal \citep{DBLP:journals/tcbb/BaiYWZWJR023/reqa_bioasq}.

Typically, a PLM well-suited to a dataset is expected to generate well-distributed sentence embeddings, where matched pairs exhibit high matching scores and vice versa, even before fine-tuning \citep{DBLP:conf/emnlp/GaoYC21/simcse}. 
Hence, the expected rank of documents relevant to queries resulted from initial model can serve as explicit and consistent indicators of model fine-tuning performance in text ranking, in contrast to the classification-oriented TE methods.
However, a challenge arises from the anisotropy that usually exists in PLMs from small size to large size, manifesting as entangled feature dimensions and a biased distribution of sentence embeddings \citep{DBLP:conf/coling/LiuZLFX024/llm_anistropy}. 
In this case, even if two sentences are very similar in semantics, their vector representations may be far apart or exhibit directional differences.
As a consequence, it leads to inaccurate document rank \citep{DBLP:journals/corr/abs-2103-15316/whitening-bert} and hinders the alignment of expected rank scores with ranking transferability.
In solving the anisotropy problem, previous research has proposed isotropization methods such as BERT-flow \citep{DBLP:conf/emnlp/LiZHWYL20/bert-flow}, BERT-whitening \citep{DBLP:journals/corr/abs-2103-15316/whitening-bert}, and SimCSE \citep{DBLP:conf/emnlp/GaoYC21/simcse}, aiming to transform sentence embeddings into an isotropic space.
Considering BERT-flow and SimCSE both need careful training \citep{DBLP:conf/emnlp/GaoYC21/simcse}, we select BERT-whitening in this work since it only involves efficiently whitening the sentence embeddings.
However, the isotropization overlooks the training dynamics involved in adapting model to the text ranking task \citep{DBLP:journals/ipm/SasakiHSI23}, hindering the full expression of the fine-tuning ranking capabilities.
To address this issue, we introduce an adaptive scaling mechanism after obtaining isotropic sentence embeddings, aiming to simulate the training dynamics by a simple but effective scaling operation, whose scaling weight can be efficiently solved by the least square method \citep{DBLP:journals/jcam/Ding23/least_square}.
Then, our approach allows for a more precise reflection of the model's true transferability through improved expected rank scores.

In summary, we present a promising alternative to human intuition for MS in text ranking: the \textbf{A}dapt\textbf{i}ve \textbf{R}anking \textbf{T}ransferability (\textbf{AiRTran}). 
In this approach, the expected rank is computed and further enhanced by the proposed \textbf{Ada}ptive \textbf{Iso}tropization (\textbf{AdaIso}) to estimate ranking transferability. 
On five widely studied datasets with two challenges pools of small and large candidate PLMs, covering various text ranking scenarios and dual-encoder backbones, our method shows superior efficiency compared to brute-force fine-tuning and show promising performance compared to previous classification-oriented TE methods, human intuition, and ChatGPT.

\section{Related Work}

Early text ranking methods were primarily based on bag-of-words functions \citep{DBLP:journals/ipm/Aizawa03/TF-IDF,DBLP:journals/ftir/RobertsonZ09/BM25}, which fall short of comprehensively understanding semantic relevance.
In recent years, with the advancements in PLMs, dual-encoder has demonstrated superiority in capturing semantic interaction that goes beyond simple word overlap \citep{zhao2024dense}.
Prior research on dual-encoder has predominantly focused on improving aspects such as sentence representations \citep{DBLP:journals/corr/abs-2104-08253/condenser,DBLP:conf/coling/TangWY22,DBLP:conf/aaai/0002MLLWH23}, negative document sampling \citep{DBLP:conf/iclr/XiongXLTLBAO21/ance,DBLP:conf/sigir/FormalLPC22,DBLP:conf/www/ChenCJYDH23}, matching score computation \citep{DBLP:journals/tacl/KhattabPZ21,DBLP:conf/sigir/LassanceMPC22,DBLP:journals/tweb/WangMTO23}, and supervisory signals construction \citep{DBLP:conf/emnlp/WangBWZRJWX21/ENDX,DBLP:conf/sigir/ZengZV22,DBLP:conf/www/LinGLZLD0LJMD23}, etc.

Unlike the above research focus on improving the network structure or training approach, our work explores selecting the most powerful PLM to achieve superior fine-tuning performance.
The most direct approaches to MS involve either exhaustively trying all available models \citep{DBLP:conf/acl/NiACMHCY22/sentence-t5}, or training a performance predictor to estimate how well a given model will perform on a dataset \citep{DBLP:journals/neurips/model_spider,DBLP:conf/nips/MengSPJZQL23}, while they are less practical due to the large demands for model training.
To address this issue, several TE methods have been developed for efficient MS, primarily in classification tasks \citep{DBLP:conf/emnlp/BaiZLHXLR23/emnlp_survey}.
These methods, when provided with model-encoded features and corresponding labels, quantify the degree of compatibility between them, treating this as a measure of transferability. 
Subsequently, they rank candidate models based on their transferability scores to identify the best-performing model.
Some of these approaches are grounded in the assumption that an effective model should produce features with a high degree of class separability \citep{DBLP:conf/iclr/PuigcerverRMRPG21/knn,DBLP:conf/cvpr/Kumari0SZ22/logistic,DBLP:conf/cvpr/PandyAUFM22/gbc,DBLP:conf/iccv/XuK23/tmi}.
Another line of research focuses on the approximation of likelihood from features to labels \citep{DBLP:conf/icip/BaoLHZZZG19/hscore,DBLP:conf/icml/YouLWL21/logme,DBLP:conf/cvpr/PandyAUFM22/gbc,DBLP:conf/icml/HuangHRY022/transrate,DBLP:conf/eccv/ShaoZGZYWSL22/sfda,DBLP:conf/eccv/DingCLCS22/pactran}.
Inspired by these remarkable works, this paper focuses on the selection of text ranking model which is an underexplored and challenging area.

\section{Preliminaries}
Before delving into the details of our proposed method, we begin by providing foundational knowledge, including the problem definition, dual-encoder model, and an examination of the challenges that previous TE methods have faced.

\subsection{Problem Definition}

\label{subsec: definition}
In the context of MS for text ranking, we are presented with a pool of $M$ candidate PLMs, denoted as $\{\phi_i\}_{i=1}^{M}$, and a dataset $\mathcal{D} = \{(q_i, d_i, y_i)\}_{i=1}^{N}$ containing $N$ pairs of (query $q_i$, document $d_i$), where the label $y_i$ is 1 if $d_i$ is relevant to $q_i$ and 0 otherwise.
To evaluate the performance of MS, we are also provided with the fine-tuning performances $\{T_i(\mathcal{D})\}_{i=1}^{M}$ of all candidate PLMs on $\mathcal{D}$  (e.g., the R@10 score). 
The goal of MS is to assign scores to all candidate PLMs, denoted as $\{S_i(\mathcal{D})\}_{i=1}^{M}$, which can well correlate with $\{T_i(\mathcal{D})\}_{i=1}^{M}$, allowing us to select the best-performing model.

\subsection{Dual-encoder}
\label{subsection: dual-encoder}
The dual-encoder is commonly employed to rank desired results from large-scale candidate documents, which consists of PLM-based query and document encoders.
The query $q$ and document $d$ are mapped into sentence embeddings by conducting mean pooling on the last layer's outputs of corresponding encoders ($e_q=\phi(q)$ and $e_d=\phi(d)$), where $e_q\in\mathbb{R}^{1\times D}$, $e_d\in\mathbb{R}^{1\times D}$, and $D$ represents the dimension of the embeddings. 
Subsequently, the matching score between the query and the document is computed by the dot-product\footnote{There are alternatives such as cosine and Euclidean distance, we select dot-product due to its ease of use.} between their sentence embeddings ($e_q e_d^T$).

For each query, the training objective of the dual-encoder is to ensure that its matching scores for relevant pairs are higher than those for irrelevant pairs. 
To achieve this goal, the probability of the relevant document $d^+$ is typically optimized:
\begin{equation}\label{equ: dual-encoder_loss}
\begin{split}
    p(d^{+}|&q,d^{+},\mathcal{I}_q)\\&= \frac{\exp(e_qe_{d^{+}}^{T})}{\exp(e_qe_{d^{+}}^{T}) + \sum\limits_{d^{-}\in \mathcal{I}_q} \exp(e_qe_{d^{-}}^{T})}
\end{split}
\end{equation}
where $d^+$ is from $\mathcal{R}_q$ that is the set of documents relevant to $q$, $d^-$ is the document irrelevant to $q$, and $\mathcal{I}_q$ is the set of documents irrelevant to $q$.

\subsection{Classification-oriented TE Methods}
Previous research has been devoted to MS for classification tasks.
Given the PLM-encoded feature $f$ and label $y$, these methods estimate the expected log conditional probabilities, denoted as $\mathbb{E}[{\rm log}(p(y|f))]$. 
This objective aligns with the training objective of the classification, establishing consistency between the estimated transferability and model fine-tuning performance.

However, text ranking involves a more intricate training objective, and merely estimating $\mathbb{E}[{\rm log}(p(y|e_q,e_d))]$ may not adequately reflect the model ranking capabilities.
To illustrate this point, let's consider a scenario where we have a query $q$ with one relevant document $d^+$ and two irrelevant documents $d^{-}_{1}$ and $d^{-}_{2}$. 
Suppose the corresponding matching probabilities from $\phi_1$ for $d^+$, $d^{-}_{1}$, and $d^{-}_{2}$ are 0.5, 0.45, and 0.45, respectively. In this case, $\mathbb{E}[{\rm log}(p(y|q,d))]$ is -0.82, and the rank of $d^+$ is 1.
Now, consider another PLM, $\phi_2$, where the matching probabilities for $d^+$, $d^{-}_{1}$, and $d^{-}_{2}$ are 0.6, 0.65, and 0.1. In this case, the expectation is -0.72, but the rank of $d^+$ is 2.
Interestingly, while $\phi_2$ has a higher transferability score, it exhibits poorer ranking ability than $\phi_1$, underscoring the necessity of estimating transferability in a manner that is more aligned with text ranking performance.

\begin{figure*}[t]
\centering
\includegraphics[width=2.1\columnwidth]{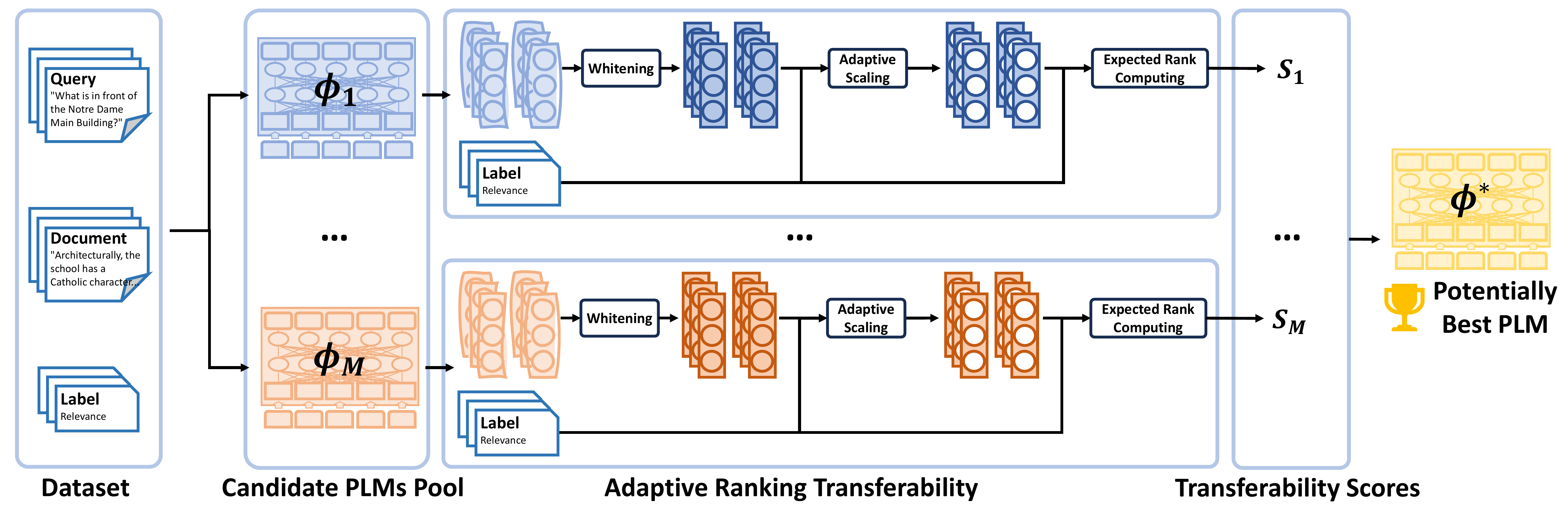}
\caption{This is the pipeline of model selection in text ranking using AiRTran. First, the queries and documents are encoded to sentence embeddings by each candidate model $\phi$. Then, the raw embeddings are transformed by whitening and adaptive scaling sequentially. Finally, the transformed embeddings coupled with labels are used to compute the expected rank as transferability, resulting in the selection of the best-performing model.}
\label{fig: overview}
\end{figure*}

\section{Methodology}

As illustrated in Figure \ref{fig: overview}, we introduce the AiRTran approach, motivated by the observation that expected rank score is inherently well-aligned with text ranking performance. 
To further improve this alignment, we propose AdaIso to mitigate the anisotropy problem and adapt raw sentence embeddings to the downstream text ranking task. 
As a result, the resultant expected rank effectively captures the true ranking transferability of the model.
In the following sections, we will first describe how to compute expected rank and then elaborate on how AdaIso further achieves the enhancement.

\subsection{Expected Rank as Transferability}
\label{subsec: ranking_metric}

The computation of expected rank score necessitates the determination of matching scores between queries and candidate documents. 
Fortunately, the dual-encoder can readily perform inference once initialized by a PLM, allowing for the direct acquisition of matching scores and resultant expected rank score. 
Consequently, the computation incurs minimal cost, merely involving a simple forward propagation process of the PLM on the dataset.
After obtaining sentence embeddings and corresponding matching scores, we can further compute the ranks of documents relevant to queries and also the expected rank (as Eq. \ref{equ: expected_rank}):
\begin{equation}
    \label{equ: expected_rank}
    S(\mathcal{D}) = \mathbb{E}_{q\sim p(q)}[\mathbb{E}_{d_+ \in\mathcal{R}_q,d_- \in\mathcal{I}_q}[\frac{1}{{\rm rank}(d_+)}]]
\end{equation}

\subsection{Adaptative Isotropization}

Due to the inherent disparity between the pre-training task and the text ranking task, sentence embeddings encoded by PLMs often exhibit anisotropy issue that renders them unsuitable for direct use in text ranking.
Anisotropy manifests as severe entanglement among the feature dimensions and distortion of the sentence embeddings' distribution, leading to matching scores that inadequately capture sentence semantics \citep{DBLP:conf/emnlp/LiZHWYL20/bert-flow}.
When working with these perturbed sentence embeddings, the computed expected rank score fails to genuinely reflect the model transferability.

Numerous approaches have been introduced to address this issue, such as BERT-flow \citep{DBLP:conf/emnlp/LiZHWYL20/bert-flow}, SimCSE \citep{DBLP:conf/emnlp/GaoYC21/simcse}, and BERT-whitening \citep{DBLP:journals/corr/abs-2103-15316/whitening-bert}.
BERT-flow and SimCSE both necessitate careful model training, where BERT-flow involves the training of a flow-based calibration model, and SimCSE optimizes the contrastive objective to achieve isotropization. 
However, these methods run counter to the objective of efficient MS, and their training is also unstable.
Hence, in this work, we opt for BERT-whitening as our primary isotropization method, which exclusively post-processes sentence embeddings through an efficient whitening operation, bypassing the need for model training \citep{DBLP:conf/emnlp/HuangTZLSGJD21}.

In BERT-whitening, the mean sentence embedding $\Vec{\mu}\in\mathbb{R}^{1\times D}$ (Eq. \ref{equ: mean}) and the covariance matrix $\Sigma\in\mathbb{R}^{D\times D}$ (Eq. \ref{equ: cov}) are initially estimated based on all sentence embeddings $E\in\mathbb{R}^{2N\times D}$:
\begin{equation}\label{equ: mean}
    \Vec{\mu} = \frac{1}{2N}\Vec{1}_{2N}^T E
\end{equation}
\begin{equation}\label{equ: cov}
\Sigma = \frac{1}{2N-1}(E-\Vec{1}_{2N}\Vec{\mu})^{T}(E-\Vec{1}_{2N}\Vec{\mu})+\epsilon I_D
\end{equation}
where $\Vec{1}_{2N}\in\mathbb{R}^{2N\times 1}$ is the $2N$-dimensional ones vector, $I_D\in \mathbb{R}^{D\times D}$ is the identity matrix, and $\epsilon\textgreater 0$ is a small positive number to prevent a singular $\Sigma$.
Then the sentence embeddings are centered and transformed as Eq. \ref{equ: whitening_transform}:
\begin{equation}\label{equ: whitening_transform}
\begin{split}
    \hat{E} = (&E-\Vec{1}_{2N}\Vec{\mu})U\sqrt{\Lambda^{-1}}
\end{split}
\end{equation}
where $\Sigma =U\Lambda U^{T}$, $\hat{E}$ is the whitened embeddings.

However, isotropization does not account for the adaptation of sentence embeddings to the downstream task, which restricts its ability to accurately reveal the true ranking performance of the model.
To address this limitation, we propose an adaptive scaling mechanism partially simulating the task adaptation by further scaling the isotropic sentence embeddings, expressed as $\hat{E}\odot (\Vec{1}_{2N}\Vec{\gamma}^T)$, where $\odot$ denotes the Hadamard product, and $\Vec{\gamma}\in \mathbb{R}^{D\times 1}$ represents the scaling weight.
The crucial question is how to determine the optimal $\Vec{\gamma}^*$. Fortunately, this weight vector can be derived by solving an ordinary least squares problem that minimizes the squared difference between predicted matching scores and ground truth labels.
To begin, given the whitened sentence embeddings for all queries ($\hat{E}_q\in \mathbb{R}^{N\times D}$) and for all documents ($\hat{E}_d\in \mathbb{R}^{N\times D}$), the predicted matching scores $\hat{Y}\in \mathbb{R}^{N\times 1}$ can be calculated as:
\begin{equation}\label{equ: prediction}
\begin{split}
    \hat{Y}&=(\hat{E}_q\odot(\Vec{1}_N\Vec{\gamma}^T))\odot(\hat{E}_d\odot(\Vec{1}_N\Vec{\gamma}^T))\Vec{1}_D\\
    &=\Vec{1}_N(\Vec{\gamma}^2)^T\odot\hat{E}_q\odot\hat{E}_d\Vec{1}_D\\
     &=(\hat{E}_q\odot\hat{E}_d)\Vec{\gamma}^2
\end{split}
\end{equation}
where $\Vec{1}_{N}\in\mathbb{R}^{N\times 1}$ and $\Vec{1}_{D}\in\mathbb{R}^{D\times 1}$.
Next, given all ground truth labels $Y\in\mathbb{R}^{N\times 1}$, the squared difference between $\hat{Y}$ and $Y$ is:
\begin{equation}\label{equ: squared_difference}
\begin{split}
    \mathcal{L}(\Vec{\gamma})&=(\hat{Y}-Y)^T(\hat{Y}-Y)\\
    &=\hat{Y}^T\hat{Y}-2\hat{Y}^TY+Y^TY\\
    &=(\Vec{\gamma}^2)^T\hat{E}_m^T\hat{E}_m\Vec{\gamma}^2-2(\Vec{\gamma}^2)^T\hat{E}_m^TY\\&\,\,\,\,\,\,+Y^TY
\end{split}
\end{equation}
where $\hat{E}_m=\hat{E}_q\odot\hat{E}_d$.
Then, by minimizing $\mathcal{L}(\Vec{\gamma})$, we can obtain the optimal $\Vec{\gamma}^*$:
\begin{equation}\label{equ: opti_gamma}
\Vec{\gamma}^*=\underset{\Vec{\gamma}}{\arg\min}\mathcal{L}(\Vec{\gamma})
\end{equation}
This can be efficiently solved by setting the partial derivative equal to zero:
\begin{equation}\label{equ: partial_derivative}
    \frac{\partial \mathcal{L}(\Vec{\gamma})}{\partial \Vec{\gamma}^2}=2\hat{E}_m^T\hat{E}_m\Vec{\gamma}^2 -2\hat{E}_m^TY=0
\end{equation}
\begin{equation}\label{equ: gamma}
    \Vec{\gamma}^*=\sqrt{(\hat{E}_m^T\hat{E}_m)^{-1}\hat{E}_m^TY}
\end{equation}
Ultimately, utilizing the sentence embeddings $\hat{E}\odot (\Vec{1}_{2N}\Vec{\gamma}^{*T})$ enhanced by adaptive isotropization, we can achieve more precise matching scores, which in turn lead to the computation of improved expected rank scores $\{S_i(\mathcal{D})\}_{i=1}^{M}$.

\section{Experiments}

\subsection{Datasets}
To validate the AiRTran, the experiments are conducted on five datasets: SQuAD \citep{DBLP:conf/emnlp/RajpurkarZLL16/squad}, NQ \citep{DBLP:journals/tacl/KwiatkowskiPRCP19/nq}, BioASQ \citep{DBLP:journals/tcbb/BaiYWZWJR023/reqa_bioasq}, SciFact \citep{DBLP:conf/emnlp/WaddenLLWZCH20/scifact}, and MuTual \citep{DBLP:conf/acl/CuiWLZZ20/mutual}, covering typical text ranking scenarios (For dataset details, see Appendix \ref{app: datasets}).
To conduct TE, we randomly sample 1,000 queries from the training set of each dataset.
The irrelevant documents for each query are then randomly sampled, since constructing irrelevant documents by human annotation or other sophisticated methods can be time-consuming, while this simple strategy also introduces a variable number of samples.
Increasing the number of irrelevant documents usually creates a more challenging scenario, effectively highlighting differences in the ranking abilities of models. 
As a result, our experiments are carried out with varying sizes of candidate documents (consisting of one relevant document and multiple irrelevant documents) for each query, ranging from 2 to 10.

\subsection{Candidate Model Pools}
We have curated two model pools from Hugging Face's model hub \citep{DBLP:conf/icse/JiangSHSSLTD23}, consisting of 25 small PLMs (11M\textasciitilde 140M parameters) and 25 large PLMs (1B\textasciitilde 8B parameters), respectively. 
We fully fine-tuned the small PLMs and parameter-efficiently fine-tuned the large ones to obtain their fine-tuning results. 
Each fine-tuning is conducted using 5 different random seeds, and the results are averaged. 
Subsequently, the best fine-tuning performances are recorded to form $\{T_i(\mathcal{D})\}_{i=1}^{M}$ (We record R@10 for SQuAD and NQ since they are text retrieval tasks that focus on recall performance, and record P@1 for BioASQ, SciFact, and MuTual since they are text matching tasks that focus on matching accuracy.) The training details and fine-tuning results are presented in Appendix \ref{app: ft_results}.

\begin{table*}[!htpb]
    \centering
    \setlength{\tabcolsep}{4pt}{
    \begin{tabular}{lcccccccccc}
        \toprule
        \multirow{2}{*}{\textbf{Methods}} & \multicolumn{2}{c}{\textbf{SQuAD}} & \multicolumn{2}{c}{\textbf{NQ}} & 
        \multicolumn{2}{c}{\textbf{BioASQ}} & \multicolumn{2}{c}{\textbf{SciFact}} & \multicolumn{2}{c}{\textbf{MuTual}}  \\
        \cmidrule(lr){2-3}\cmidrule(lr){4-5}\cmidrule(lr){6-7}\cmidrule(lr){8-9}\cmidrule(lr){10-11}
        ~ & $\bm{\tau_{\rm{Small}}}$ & $\bm{\tau_{\rm{Large}}}$ & $\bm{\tau_{\rm{Small}}}$ & $\bm{\tau_{\rm{Large}}}$ &$\bm{\tau_{\rm{Small}}}$ & $\bm{\tau_{\rm{Large}}}$ & $\bm{\tau_{\rm{Small}}}$ & $\bm{\tau_{\rm{Large}}}$ & $\bm{\tau_{\rm{Small}}}$ & $\bm{\tau_{\rm{Large}}}$ \\
        \midrule
TMI & 0.252 & 0.332 & 0.275 & 0.379 & 0.143 & 0.356 & 0.496 & 0.439 & 0.255 & 0.248 \\
GBC & 0.375 & 0.513 & 0.529 & 0.496 & 0.180 & 0.555 & 0.416 & 0.541 & 0.080 & 0.383 \\
TransRate & 0.592 & 0.680 & 0.327 & 0.408 & 0.313 & 0.315 & 0.147 & 0.520 & 0.172 & 0.180 \\
$\mathcal{N}$LEEP & 0.552 & 0.436 & 0.516 & 0.456 & 0.303 & 0.363 & 0.559 & 0.328 & 0.112 & 0.448 \\
LinearProxy & 0.454 & 0.477 & 0.359 & 0.546 & 0.339 & 0.394 & 0.629 & 0.551 & 0.229 & 0.577 \\
SFDA & 0.639 & 0.597 & 0.623 & 0.593 & 0.441 & 0.411 & 0.693 & 0.498 & 0.229 & 0.503 \\
PACTran & 0.643 & 0.656 & 0.661 & 0.668 & 0.395 & 0.567 & 0.681 & 0.571 & 0.164 & 0.564 \\
H-score & 0.629 & 0.663 & 0.608 & 0.627 & 0.435 & 0.596 & 0.696 & \textbf{\underline{0.617}} & 0.223 & 0.513 \\
LogME & 0.645 & \textbf{\underline{0.720}} & 0.621 & 0.660 & 0.417 & 0.583 & 0.695 & 0.549 & 0.125 & 0.528 \\

AiRTran & \textbf{\underline{0.667}} & 0.706 & \textbf{\underline{0.672}} & \textbf{\underline{0.674}} & \textbf{\underline{0.508}} & \textbf{\underline{0.648}} & \textbf{\underline{0.785}} & 0.600 & \textbf{\underline{0.351}} & \textbf{\underline{0.583}} \\

        \bottomrule
        
    \end{tabular}}
    \caption{The best $\tau$ of all TE methods over different document sizes, where the highest scores are underlined.}
    \label{tab: main_results}
\end{table*}

\begin{figure*}[t]
\centering
\includegraphics[width=2.05\columnwidth]{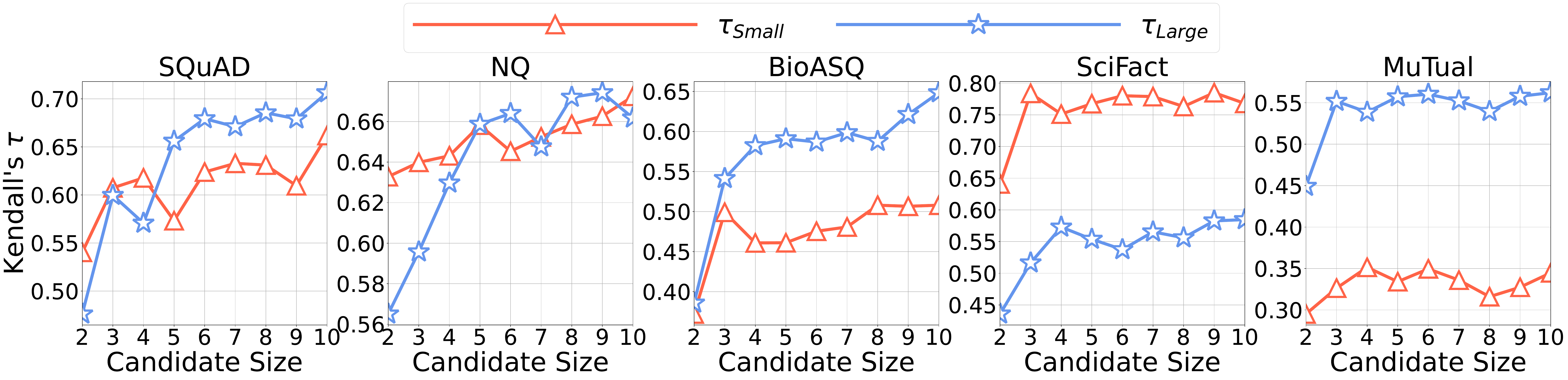}
\caption{The performance variations of AiRTran over different document sizes.}
\label{fig: airtran_variation}
\end{figure*}

\subsection{Evaluation Metric}
Given the target of MS, when $S_i$ is higher than $S_j$, the corresponding $T_i$ is expected to be higher than $T_j$.
To evaluate TE approaches, we employ Kendall's $\tau$ to measure the degree of agreement between $\{S_i(\mathcal{D})\}_{i=1}^{M}$ and $\{T_i(\mathcal{D})\}_{i=1}^{M}$.
It is particularly valuable when working with ordinal or ranked data, without any assumptions about the underlying data distribution.
The formula for calculating Kendall's $\tau$ is as follows:
\begin{equation}\label{equ: kendall_tau}
\tau=\frac{2\sum_{1\leq i\le j\leq M}\mathbb{I}(T_i-T_j)\mathbb{I}(S_i-S_j))}{M(M-1)}
\end{equation}
where $\tau\in[-1,1]$; the indicator function $\mathbb{I}$ returns 1 for positive number and return -1 otherwise.
For each TE method, we compute $\tau_{\rm{Small}}$, $\tau_{\rm{Large}}$ that denote the Kendall's $\tau$ computed when $T$ corresponds to the fine-tuning results of small model pool and large model pool, respectively.

\subsection{Compared Approaches}
We compare the following competitive TE methods which have proven effective when facing classification tasks: TMI \citep{DBLP:conf/iccv/XuK23/tmi}, GBC \citep{DBLP:conf/cvpr/PandyAUFM22/gbc}, TransRate \citep{DBLP:conf/icml/HuangHRY022/transrate}, $\mathcal{N}$LEEP \citep{DBLP:conf/cvpr/LiJSZGWG21/nleep}, LinearProxy \citep{DBLP:conf/cvpr/Kumari0SZ22/logistic}, SFDA \citep{DBLP:conf/eccv/ShaoZGZYWSL22/sfda}, PACTran \citep{DBLP:conf/eccv/DingCLCS22/pactran}, H-score \citep{DBLP:conf/icip/BaoLHZZZG19/hscore}, LogME \citep{DBLP:conf/icml/YouLWL21/logme}, whose details are presented in Appendix \ref{app: methods}.
We run each method using 5 random seeds and report averaged Kendall's $\tau$.
Note that these methods assume each data sample corresponds to one feature vector. 
To adapt these methods for text ranking, where each sample consists of a pair of sentence embeddings, we explored Hadamard product, element-wise addition, element-wise subtraction, and concatenation, to combine two sentence embeddings into a single feature vector. 
We found that all methods achieved their best performance when utilizing the Hadamard product operation\footnote{For implementation details, our code is available at \href{https://github.com/Ba1Jun/model-selection-AiRTran}{https://github.com/Ba1Jun/model-selection-AiRTran}.}.

\subsection{Model Selection Performance}

Table \ref{tab: main_results} presents the best results of all methods.
The anisotropy issue leads to substantial overlap between features of matched and mismatched pairs, which hinders the accurate computation of Gaussian distance and entropy, resulting in inferior performances for TMI, GBC, and TransRate.
The methods that simulate training dynamics including LinearProxy, SFDA, PACTran, and LogME exhibit strong performance.
H-score also demonstrates notable improvements, attributed to its consideration of feature redundancy in addition to the inter-class variance.
Note that $\mathcal{N}$LEEP involves the feature transformation as well, while it doesn't consider the label information, thus resulting in poor performance. 
Nevertheless, these classification-oriented methods generally trail behind AiRTran due to their lack of alignment with the text ranking ability.
With the inclusion of AdaIso which adaptively scales the isotropic sentence embeddings, AiRTran generates expected rank that strongly correlates with model fine-tuning performance. 
Notably, it achieves the highest $\tau$ in 8 out of 10 cases, providing strong validation of its effectiveness. For additional results and specific predictions of AiRTran, please see Appendix \ref{app: add_results} and \ref{app: airtran_prediction}.

\begin{table*}[!htpb]
    \centering
    \setlength{\tabcolsep}{3pt}{
    \begin{tabular}{lcccccccccc}
        \toprule
        \multirow{2}{*}{\textbf{Methods}} & \multicolumn{2}{c}{\textbf{SQuAD}} & \multicolumn{2}{c}{\textbf{NQ}} & 
        \multicolumn{2}{c}{\textbf{BioASQ}} & \multicolumn{2}{c}{\textbf{SciFact}} & \multicolumn{2}{c}{\textbf{MuTual}}  \\
        \cmidrule(lr){2-3}\cmidrule(lr){4-5}\cmidrule(lr){6-7}\cmidrule(lr){8-9}\cmidrule(lr){10-11}
        ~ & $\bm{\tau_{\rm{Small}}}$ & $\bm{\tau_{\rm{Large}}}$ & $\bm{\tau_{\rm{Small}}}$ & $\bm{\tau_{\rm{Large}}}$ &$\bm{\tau_{\rm{Small}}}$ & $\bm{\tau_{\rm{Large}}}$ & $\bm{\tau_{\rm{Small}}}$ & $\bm{\tau_{\rm{Large}}}$ & $\bm{\tau_{\rm{Small}}}$ & $\bm{\tau_{\rm{Large}}}$ \\
        \midrule

AiRTran (flow) & 0.662 & 0.630 & 0.652 & 0.538 & 0.451 & 0.602 & 0.764 & \textbf{\underline{0.663}} & 0.349 & 0.440 \\
AiRTran (simcse) & 0.635 & 0.657 & \textbf{\underline{0.676}} & 0.602 & \textbf{\underline{0.576}} & 0.606 & \textbf{\underline{0.811}} & 0.605 & 0.320 & 0.453 \\
\hline
AiRTran (whiten) & \textbf{\underline{0.667}} & \textbf{\underline{0.706}} & 0.672 & \textbf{\underline{0.674}} & 0.508 & \textbf{\underline{0.648}} & 0.785 & 0.600 & \textbf{\underline{0.351}} & \textbf{\underline{0.583}} \\
\quad \quad  w/o Iso & 0.646 & 0.419 & 0.636 & 0.449 & 0.431 & 0.377 & 0.720 & 0.505 & 0.297 & 0.448 \\
\quad \quad w/o Ada & 0.655 & 0.688 & 0.645 & 0.647 & 0.235 & 0.580 & 0.698 & 0.556 & 0.116 & 0.226 \\
\quad \quad  w/o AdaIso & 0.561 & 0.319 & 0.554 & 0.398 & 0.219 & 0.346 & 0.484 & 0.421 & 0.223 & 0.406 \\
        \bottomrule
    \end{tabular}}
    \caption{The results of AiRTran when BERT-flow (flow), SimCSE (simcse), and BERT-whitening (whiten) are employed, respectively. The results when different AdaIso components are removed are also listed.}
    \label{tab: ablation_results}
\end{table*}

\subsection{Effect of Candidate Document Size}

As depicted in Figure \ref{fig: airtran_variation}, we visualize the performance variations of AiRTran,
and the visualization results of other methods are presented in Appendix \ref{app: variation}.
It is observed that TMI, TransRate, and $\mathcal{N}$LEEP show significant fluctuation.
The other methods indicate relatively regular patterns of performances with the candidate document size increases, while they behave differently when facing different model pools.
Both of the observations are attributed to the lack of alignment between their estimation score and model ranking ability.
In contrast, AiRTran generates expected rank score that can harness the benefits of including more irrelevant documents, thereby enhancing the differentiation of ranking capabilities among different models. 
Consequently, on both small model pool and large model pool, AiRTran's performance generally exhibits an upward trend as the candidate document size increases. 
This brings convenience to the determination of its hyper-parameters, the desired performance usually can be achieved when candidate document size is set to 10.

\begin{figure}[t]
\centering
\includegraphics[width=\columnwidth]{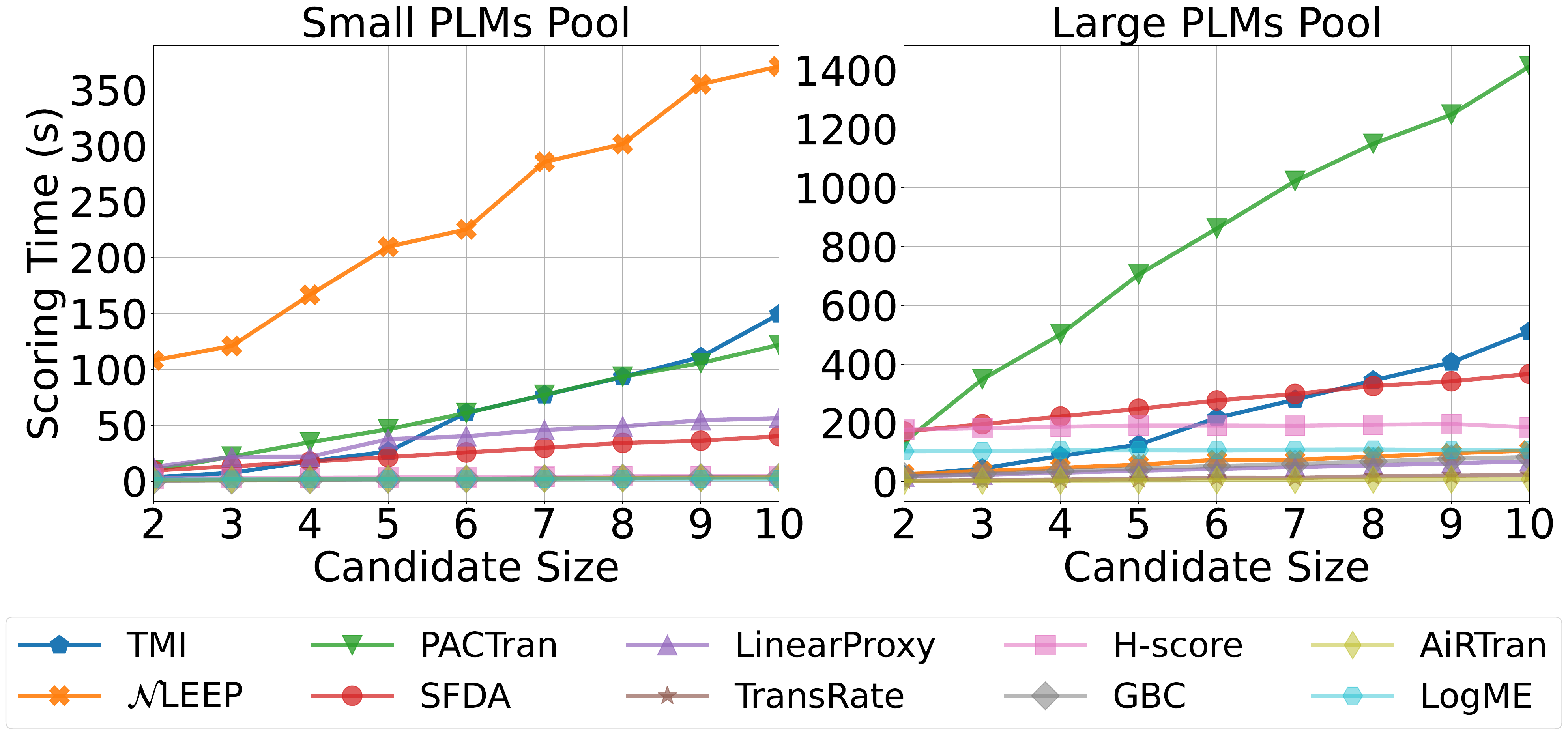}
\caption{This is the comparison between the time consumption of all methods as the size of candidate documents grows. Note that the encoding time for dataset is not included, since it is shared by all methods.}
\label{fig: time}
\end{figure}

\subsection{Scoring Time Comparison}

The scoring time, i.e., the time required for each TE method to score the transferabilities of all models given the model-encoded features, serves as a reflection of the efficiency of model selection.
In Figure \ref{fig: time}, we have visualized the trends in scoring time for all TE methods on two pools as the candidate document size increases.
Broadly, there exists a linear relationship between the scoring time of all TE methods and the size of the candidate document set.
Though the scoring time for all TE methods is significantly lower than the time required for the brute-force fine-tuning (In our case, fine-tuning all PLMs on all datasets took nearly three months using a single NVIDIA GeForce RTX 3090 24G.), the proposed AiRTran consistently operates with a fast runtime, thanks to the efficient whitening operation and the solution to the scaling weight.

\subsection{Exploration of AdaIso}
Though AiRTran is agnostic to isotropization method, we select whitening due to its efficacy.
To verify this choice, we also instantiate AiRTran by BERT-flow and SimCSE, i.e., AiRTran (flow) and AiRTran (simcse), and compare them with AiRTran (whiten).
As shown in Table \ref{tab: ablation_results}, since the training of BERT-flow and SimCSE both need to be carefully tuned, the resultant AiRTran performances are unstable. 
Specifically, AiRTran (flow) performs best only in the SciFact with the large model. 
Though AiRTran (simcse) performs well in several datasets (e.g., NQ with small model and SciFact with small model), but overall, it is slightly outperformed by the whitening variant.
In contrast, AiRTran (whiten) consistently shows the highest performance across most cases, generally being the most effective instantiation.
Moreover, AiRTran (whiten) requires only seconds of CPU runtime to perform the whitening, whereas the other methods require costly GPU resources.
We also conducted an ablation study for AiRTran (whiten) by removing isotropization (w/o Iso), adaptive scaling (w/o Ada), and full AdaIso (w/o AdaIso), 
the results revealed that both adaptive scaling and isotropization play a crucial role, where removing any of them led to a significant drop.
Additionally, our observations highlight that isotropization does not consistently yield improvements, 
e.g., ``w/o Ada'' performed worse than ``w/o AdaIso'' on MuTual, 
primarily due to its unsupervised nature, underscoring the necessity of adaptive scaling.

\begin{table}[!tpb]
    \centering
    \setlength{\tabcolsep}{3pt}{
    \begin{tabular}{lcccc}
        \toprule
        \multirow{2}{*}{\textbf{Methods}} & \multicolumn{2}{c}{\textbf{SQuAD}} & \multicolumn{2}{c}{\textbf{BioASQ}}  \\
        \cmidrule(lr){2-3}\cmidrule(lr){4-5}
        ~ & $\bm{\tau_{\rm{Small}}}$ & $\bm{\tau_{\rm{Large}}}$ & $\bm{\tau_{\rm{Small}}}$ & $\bm{\tau_{\rm{Large}}}$ \\
        \midrule
$\mathcal{Q}$Tran & 0.437 & 0.393 & 0.267 & 0.393 \\
\quad\quad  w Ada & \textbf{\underline{0.600}} & 0.580 & 0.347 & 0.573 \\
\quad\quad  w Iso & 0.467 & 0.447 & 0.312 & 0.520 \\
\quad\quad  w AdaIso & 0.567 & \textbf{\underline{0.673}} & \textbf{\underline{0.450}} & \textbf{\underline{0.580}} \\
        \bottomrule
    \end{tabular}}
    \caption{The Kendall's $\tau$ performance of $\mathcal{Q}$Tran when different components of AdaIso are employed.}
    \label{tab: align_uniform_results}
\end{table}

\subsection{Relation to Alignment and Uniformity}
We further use the following properties to investigate the inner workings of AdaIso:
(1) Alignment: It quantifies the expected affinity between sentence embeddings of paired sentences.
(2) Uniformity: It measures how uniformly the embeddings of different sentences are distributed \citep{DBLP:conf/icml/0001I20/uniform_align}.
These two properties align well with the objective that matched pairs should remain close in embedding space, while embeddings for random instances should be widely scattered in the context of text ranking.
As Eq. \ref{equ: align_uniform}, we combine them to form a quality score, denoted as $\mathcal{Q}$.
\begin{equation}\label{equ: align_uniform}
\mathcal{Q}= \underbrace{\mathop{\mathbb{E}}\limits_{(q,d^+)\sim P_{pos}}e_qe_{d^+}^T}_{Alignment}+\underbrace{\mathop{\mathbb{E}}\limits_{(x,x')\sim P_{data}}-e_x e_{x'}^T}_{Uniformity}
\end{equation}
Intuitively, a powerful model should exhibit high alignment and uniformity scores. 
Consequently, $\mathcal{Q}$ should exhibit a strong correlation with the model's fine-tuning results, serving as a valuable indicator for transferability, i.e., Quality score as Transferability ($\mathcal{Q}$Tran). 
However, this correlation is challenging to fully establish when the anisotropy problem persists, and training dynamics are not considered.
Fortunately, the proposed AdaIso can solve this problem. 
As demonstrated in Table \ref{tab: align_uniform_results}, it is observed that $\mathcal{Q}$Tran computed on raw sentence embeddings only exhibits a weak correlation with model fine-tuning results. 
After whitening isotropization and adaptive scaling are conducted on the raw sentence embeddings, the correlations improve significantly, facilitating more accurate MS.
We also observed that when isotropization is conducted, the further employment of adaptive scaling doesn't always bring improvement (i.e., $\tau_{Small}$ in SQuAD), probably because the sentence embeddings lose too much original information after whitening and adaptive scaling weight is over-fitted on such perturbed features.

\subsection{Comparison with Human and ChatGPT}
\begin{figure}[t]
\centering
\includegraphics[width=\columnwidth]{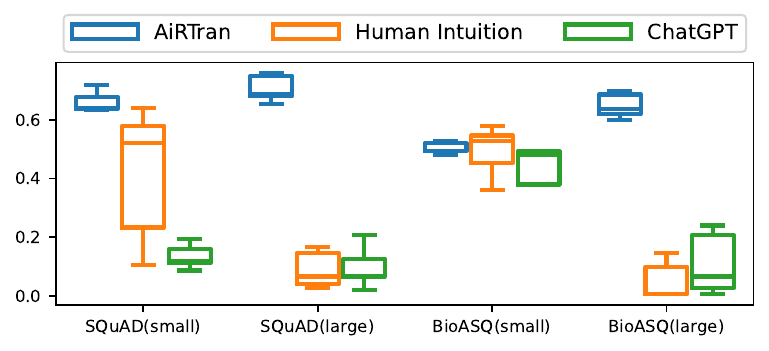}
\caption{The comparison of Kendall’s $\tau$ between AiRTran, human intuition, and ChatGPT.}
\label{fig: human}
\end{figure}

Currently, the most prevalent method for MS often relies on human intuition or, alternatively, prompts to ChatGPT \citep{chatgpt}. 
Although such processes can be completed in a few minutes, the corresponding performances become unreliable when facing challenging MS scenarios.
To demonstrate AiRTran's potential as a viable replacement for these methods, we conducted a comparative study among AiRTran, human intuition, and ChatGPT on SQuAD and BioASQ datasets.
For MS by human intuition, we engaged 5 NLP practitioners (doctoral and master's students specializing in NLP) who were provided with metadata about the dataset and models. Similarly, ChatGPT was guided using instructions that utilized the same information (see Appendix \ref{app: chatgpt}), and we parsed the ranking results of candidate PLMs from its responses. This process was repeated 5 times to capture the inherent variability in ChatGPT's responses.
The results, depicted in Figure \ref{fig: human}, emphasize the underperformance of human intuition and ChatGPT in terms of relevance and stability. This is due to their limited understanding of the dataset and models. In contrast, AiRTran excels in demonstrating transferability by effectively capturing the compatibility between model-encoded features and labels, thus showcasing a significant advantage.

\section{Conclusion}

Given the limitations of classification-oriented TE methods, we propose the use of AiRTran in this study. 
It demonstrates superior MS performance compared to competitive TE methods, human experts, and the ChatGPT MS agent, indicating its potential as a solution for MS in text ranking tasks. 
We hope that our work will offer text ranking practitioners valuable guidance and inspiration when selecting models for their datasets of interest in addition to their intuition.

\clearpage

\section*{Acknowledgements}
This work was supported by the Natural Science Foundation of China (No. 62377002).

\section*{Limitations}
To construct irrelevant documents for each query, this study employs a random sampling strategy. While easy to implement and efficient in generating irrelevant pairs, it often results in documents that are too easy to discriminate. In such an uncompetitive ranking scenario, both proficient and average models may perform well and achieve high transferability scores, complicating model selection. Therefore, there is a pressing need to explore more effective strategies for negative sampling in MS tasks.
Additionally, although human experts and ChatGPT exhibit lower performance compared to AiRTran, their knowledge demonstrates a sophisticated understanding of model-to-dataset transfer. However, this study does not investigate the potential effectiveness of integrating their insights as a complement to the feature-label interactions captured by TE approaches.

\bibliography{custom}

\appendix

\section{Datasets}
\label{app: datasets}
The dataset used in this research are:

\paragraph{SQuAD} The Stanford Question Answering Dataset (SQuAD) \citep{DBLP:conf/emnlp/RajpurkarZLL16/squad} is a widely-used benchmark for evaluating machine comprehension and passage retrieval. Created by researchers at Stanford University, SQuAD consists of over 100,000 questions derived from a diverse set of Wikipedia articles. Each question is paired with a specific passage from the article. 
The dataset is designed to test a model's ability to understand context and extract relevant information accurately. SQuAD has been instrumental in advancing the field of natural language processing, providing a rigorous standard for comparing the performance of various passage retrieval models. 
It has led to significant developments in neural network architectures and training techniques\footnote{\href{https://rajpurkar.github.io/SQuAD-explorer}{https://rajpurkar.github.io/SQuAD-explorer}}.

\paragraph{NQ} The Natural Questions (NQ) \citep{DBLP:journals/tacl/KwiatkowskiPRCP19/nq} dataset is a large-scale corpus created to facilitate research in question answering and passage retrieval. Developed by Google, it consists of real anonymized queries posed by users to the Google search engine, along with corresponding answers derived from Wikipedia articles. The dataset contains over 300,000 questions, with each entry featuring the original query, a long answer (typically a paragraph), and a short answer (usually a span of text). The long answers provide context, while the short answers aim to pinpoint precise information. This dataset is particularly valuable for training and evaluating models in tasks such as machine comprehension, information retrieval, and automated question answering\footnote{\href{https://ai.google.com/research/NaturalQuestions}{https://ai.google.com/research/NaturalQuestions}}.

\paragraph{BioASQ} The BioASQ \citep{DBLP:journals/tcbb/BaiYWZWJR023/reqa_bioasq} dataset is a comprehensive resource aimed at advancing the field of biomedical question answering. It is part of the larger BioASQ challenge, which focuses on creating systems capable of understanding and answering complex biomedical queries. The dataset comprises a vast collection of questions generated by biomedical experts, along with corresponding answers extracted from scientific literature, including PubMed abstracts and full-text articles. The questions range from simple factoid and list questions to more complex ones requiring detailed and precise answers. The BioASQ dataset promotes the development of advanced natural language processing models that can handle the intricacies of biomedical terminology and information. It is instrumental in pushing forward research in biomedical information retrieval, question answering, and semantic indexing. We select version of 9b which has 5,828 training question-answer pairs, 496 test queries, and 31,682 candidate answers\footnote{\href{http://participants-area.bioasq.org/datasets/}{http://participants-area.bioasq.org/datasets}}.

\paragraph{SciFact} The SciFact \citep{DBLP:conf/emnlp/WaddenLLWZCH20/scifact} dataset is a specialized corpus designed to facilitate research in scientific claim verification and fact-checking. Developed by the Allen Institute for AI, SciFact contains a collection of scientific claims and corresponding evidence sentences extracted from peer-reviewed biomedical research papers. The dataset includes over 1,400 claims, each accompanied by supporting or refuting evidence, enabling the development and evaluation of models that can assess the veracity of scientific statements. SciFact aims to address the challenges of verifying complex, technical claims and provides a valuable resource for advancing NLP models in the context of scientific discourse. This dataset is instrumental in promoting the development of AI systems capable of critical evaluation in scientific research\footnote{\href{https://leaderboard.allenai.org/scifact/submissions/public}{https://leaderboard.allenai.org/scifact/submissions/public}}.

\paragraph{MuTual} The MuTual \citep{DBLP:conf/acl/CuiWLZZ20/mutual} dataset is a benchmark designed to evaluate dialogue systems, particularly in the context of multi-turn reasoning. Created by researchers at Tsinghua University, MuTual consists of over 8,800 dialogues, which are sourced from real-world Chinese student English listening comprehension exams and then translated into English. Each dialogue includes a series of turns between participants, with a final turn containing four potential responses. The task is to choose the most appropriate response based on the preceding conversation. The MuTual dataset emphasizes the need for models to grasp context, manage dialogue coherence, and understand nuanced interactions. It serves as a critical resource for advancing dialogue systems and enhancing their capability to engage in meaningful, coherent, and contextually appropriate conversations\footnote{\href{https://nealcly.github.io/MuTual-leaderboard/}{https://nealcly.github.io/MuTual-leaderboard}}.

\section{Fine-tuning Candidate Models}
\label{app: ft_results}
In this work, we construct two candidate model pools of different model sizes to meet different computing needs, formed by 25 small PLMs and 25 large PLMs widely used.
We employ full fine-tuning and parameter-efficient fine-tuning for small PLMs and large PLMs, respectively, considering their common fine-tuning strategy.
Specifically, the training batch size is set to 32, and fine-tuning is terminated if there is no improvement in validation performance for 3 consecutive epochs, with a maximum of 10 training epochs.
We select AdamW \citep{DBLP:conf/iclr/LoshchilovH19/adamw} as optimizer, and we mainly tune the learning rate, which varies within [5e-6, 1e-5, 2e-5, 3e-5, 4e-5, 5e-5], as it significantly impacts the fine-tuning performance.
The official model names of these models presented in Hugging Face's model hub and their fine-tuning results on each dataset are listed in Tables \ref{tab: fine_tuning_results}.

\section{Details of Compared Methods}
\label{app: methods}

\paragraph{TMI} It views transferability as the generalization of a pre-trained model on a target task by measuring intra-class feature variance using conditional entropy \citep{DBLP:conf/iccv/XuK23/tmi}.

\paragraph{GBC} It models the in-class features by Gaussian distribution, then computes the inter-Gaussian distance as the inter-class overlap. Then it selects model by the assumption that smaller overlap results in greater transferability \citep{DBLP:conf/cvpr/PandyAUFM22/gbc}.

\begin{table*}[!htpb]
    \centering
    \setlength{\tabcolsep}{4pt}{
    \begin{tabular}{lcccccccccc}
        \toprule
        \multirow{2}{*}{\textbf{Methods}} & \multicolumn{2}{c}{\textbf{SQuAD}} & \multicolumn{2}{c}{\textbf{NQ}} & 
        \multicolumn{2}{c}{\textbf{BioASQ}} & \multicolumn{2}{c}{\textbf{SciFact}} & \multicolumn{2}{c}{\textbf{MuTual}}  \\
        \cmidrule(lr){2-3}\cmidrule(lr){4-5}\cmidrule(lr){6-7}\cmidrule(lr){8-9}\cmidrule(lr){10-11}
        ~ & $\bm{\tau_{\rm{Small}}}$ & $\bm{\tau_{\rm{Large}}}$ & $\bm{\tau_{\rm{Small}}}$ & $\bm{\tau_{\rm{Large}}}$ &$\bm{\tau_{\rm{Small}}}$ & $\bm{\tau_{\rm{Large}}}$ & $\bm{\tau_{\rm{Small}}}$ & $\bm{\tau_{\rm{Large}}}$ & $\bm{\tau_{\rm{Small}}}$ & $\bm{\tau_{\rm{Large}}}$ \\
        \midrule
        TMI & 14.2 & 2.0 & 5.6 & 2.0 & 6.0 & 8.0 & 7.0 & 2.0 & \textbf{\underline{10.0}} & 10.0 \\
GBC & 16.8 & \textbf{\underline{1.0}} & 4.0 & \textbf{\underline{1.0}} & 6.0 & 6.0 & 2.8 & 2.0 & 24.0 & 1.0 \\
TransRate & 24.0 & 23.0 & 23.0 & 19.6 & \textbf{\underline{2.0}} & 24.0 & 21.0 & 2.8 & 25.0 & 25.0 \\
$\mathcal{N}$LEEP & 8.6 & 10.0 & 3.4 & 4.4 & 4.6 & 5.2 & 4.4 & 3.2 & 21.6 & \textbf{\underline{1.0}} \\
LinearProxy & 2.0 & 6.2 & 3.0 & 7.6 & 14.2 & 1.4 & \textbf{\underline{1.8}} & 5.8 & 13.8 & 1.8 \\
SFDA & 3.0 & \textbf{\underline{1.0}} & 4.0 & \textbf{\underline{1.0}} & 5.0 & 22.6 & 2.4 & 2.0 & 23.2 & 4.0 \\
PACTran & 3.4 & 2.2 & 3.0 & 3.0 & 5.0 & 9.0 & 3.0 & 3.8 & 23.0 & 3.8 \\
H-score & 2.8 & 2.0 & 3.0 & 3.0 & 5.0 & 7.0 & 2.0 & 2.4 & 23.2 & 5.0 \\
LogME & 3.6 & 2.2 & 3.0 & 3.0 & 5.0 & 11.0 & 2.8 & 2.0 & 24.0 & \textbf{\underline{1.0}} \\
AiRTran & \textbf{\underline{1.2}} & 2.6 & \textbf{\underline{2.6}} & 1.8 & 6.0 & \textbf{\underline{1.2}} & 2.2 & \textbf{\underline{1.8}} & 22.8 & \textbf{\underline{1.0}} \\
        \bottomrule
        
    \end{tabular}}
    \caption{All TE methods' best mean estimated rank results of the best-performing model over different document sizes, where the highest scores are underlined.}
    \label{tab: rank_results}
\end{table*}

\paragraph{TransRate} It estimates the mutual information between model-encoded features and labels as transferability, where the challenge of mutual information estimation is overcome by resorting to coding rate \citep{DBLP:conf/icml/HuangHRY022/transrate}.

\paragraph{$\mathcal{N}$LEEP} It fits a Gaussian Mixture Model (GMM) to model-encoded features, then computes the likelihood of labels given the GMM-produced posterior cluster assignment to measure the fine-tuning performance \citep{DBLP:conf/cvpr/LiJSZGWG21/nleep}.

\paragraph{LinearProxy} It assumes a logistic classification model is on the top of frozen pre-trained model, then uses the cross-validation performance of such model as transferability \citep{DBLP:conf/cvpr/Kumari0SZ22/logistic}.

\paragraph{SFDA} It first embeds the static features into a Fisher space and refines them for better separability. Then, it uses a self-challenging mechanism to compute the log-loss as transferability, which encourages different pre-trained models to differentiate on hard examples \citep{DBLP:conf/eccv/ShaoZGZYWSL22/sfda}.

\paragraph{PACTran} It seeks an optimal yet efficient PAC-Bayesian bound to the generalization error in a transfer learning setting, and the error is based on the cross-entropy loss between the prediction and the labels, measuring the generalization gap of model \cite{DBLP:conf/eccv/DingCLCS22/pactran}.

\paragraph{H-score} Based on statistical and information-theoretic principles, it shows that the expected log-loss of using a model-encoded feature to predict the label of a given task under the probabilistic model can be characterized by an analytical expression, serving as model transferability \citep{DBLP:conf/icip/BaoLHZZZG19/hscore}.

\paragraph{LogME} It calculates the maximum evidence to measure a model's performance on a new task. The process involves extracting features from the pre-trained model, assuming a linear model for the new task, and computing the log marginal likelihood to assess feature suitability \citep{DBLP:conf/icml/YouLWL21/logme}.

\begin{table}[t]
    \centering
    \setlength{\tabcolsep}{28pt}{
    \begin{tabular}{lc}
        \toprule
        \textbf{Methods} & \textbf{AirTran} \\
        \midrule
        TMI & 1.83e-24 \\ 
        GBC & 1.31e-11 \\ 
        TransRate & 3.63e-13 \\ 
        $\mathcal{N}$LEEP & 1.26e-12 \\ 
        LinearProxy & 3.43e-09 \\ 
        SFDA & 2.44e-04 \\ 
        PACTran & 2.78e-02 \\ 
        H-score & 4.13e-02 \\ 
        LogME & 2.90e-02 \\
        \bottomrule
    \end{tabular}}
    \caption{The p-values of the t-test conducted on the results of AirTran and other model selection methods.}
    \label{tab: significance_test_1}
\end{table}

\begin{figure*}[h]
\centering
\includegraphics[width=2.05\columnwidth]{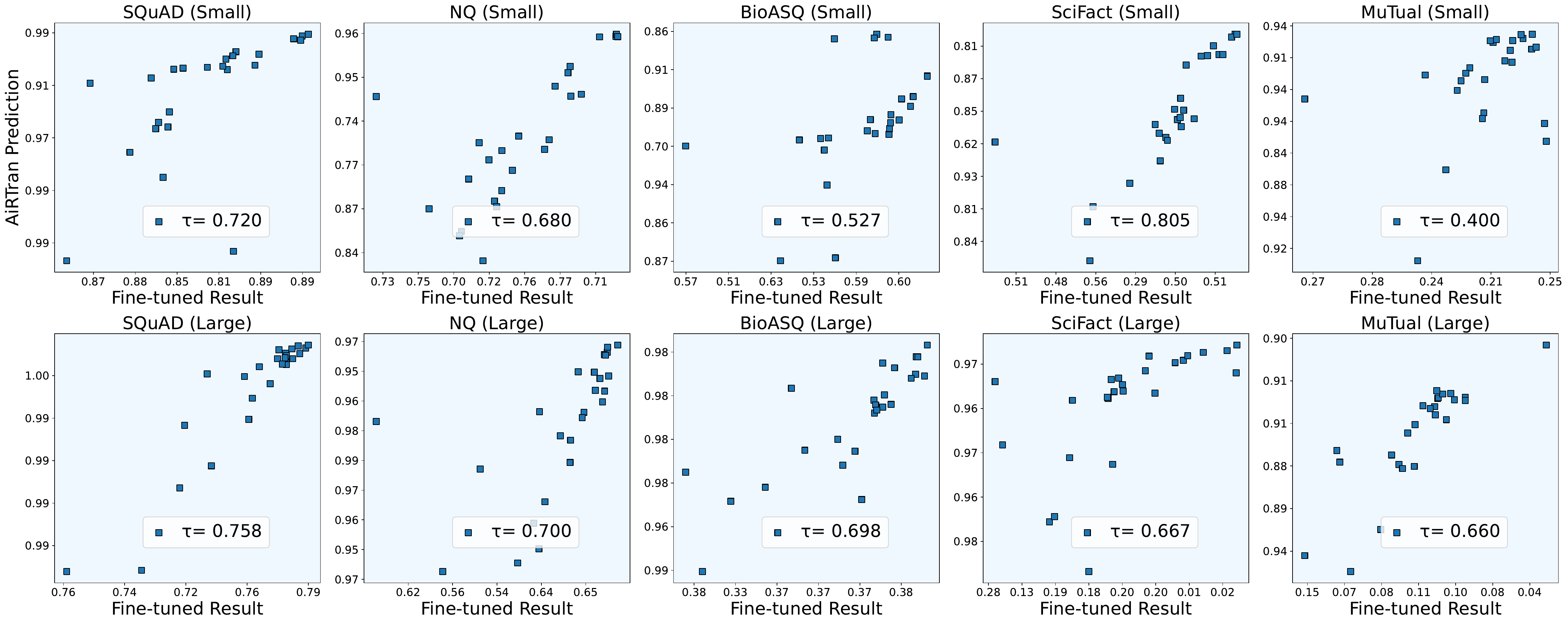}
\caption{The predictions of AiRTran against the fine-tuning results with the best Kendall's $\tau$ performance.}
\label{fig: airtran_scatter}
\end{figure*}

\section{Additional Model Selection Results}
\label{app: add_results}

To validate the improvements, we conduct significance test between the results of AiRTran and other methods. Table \ref{tab: significance_test_1} shows the p-values of the t-test for the results in Table \ref{tab: main_results}, it is observed that all p-values are less than 0.05.

We also provide all methods’ mean estimated rank results of the best-performing model. As shown in Table \ref{tab: rank_results}, we can observe that AiRTran can also rank the best model at the top in most cases, achieving 5 out of 10 best performances.

\section{The Predictions of AiRTran}
\label{app: airtran_prediction}
The best-performing predictions of AiRTran against the models' fine-tuning results on SQuAD, NQ, BioASQ, SciFact, and MuTual across runs are visualized in Figure \ref{fig: airtran_scatter}.

\section{Performance Variations of TE Methods}
\label{app: variation}
\begin{figure*}[t]
  \centering
    \subfigure[TMI]{\label{fig: variation_tmi}\includegraphics[width=2\columnwidth]{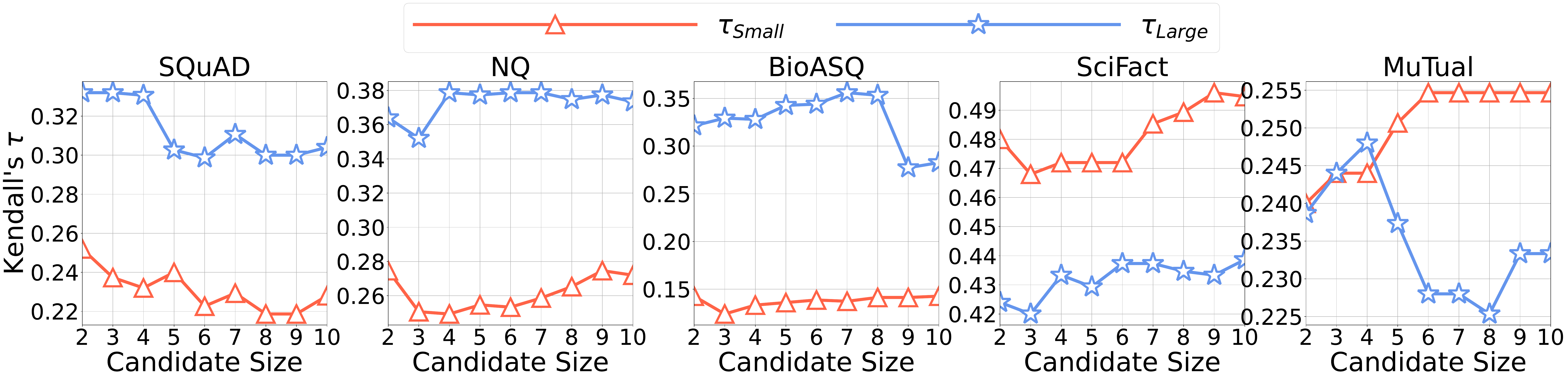}}
    \subfigure[GBC]{\label{fig: variation_gbc}\includegraphics[width=2\columnwidth]{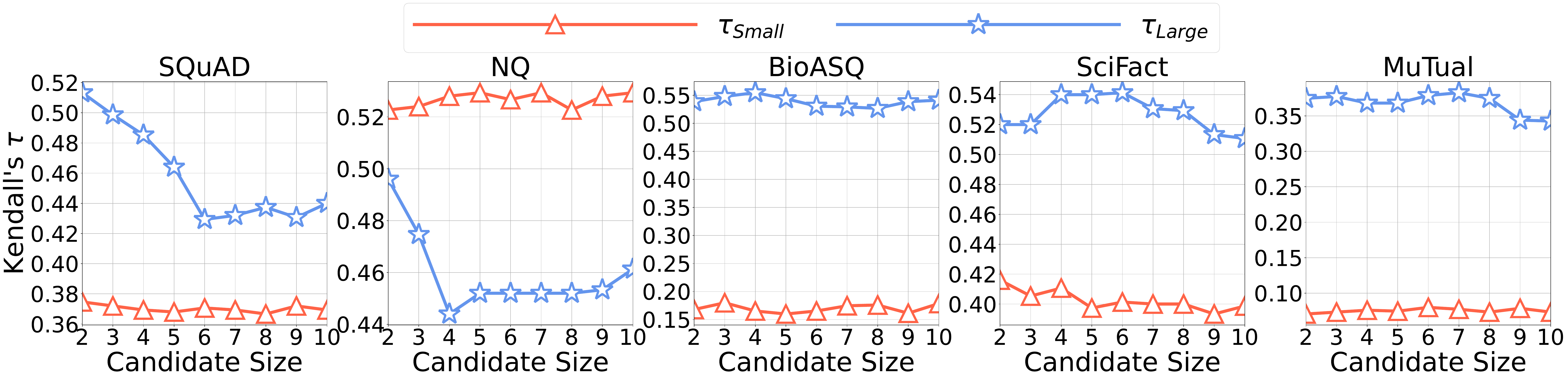}}
    \subfigure[TransRate]{\label{fig: variation_transrate}\includegraphics[width=2\columnwidth]{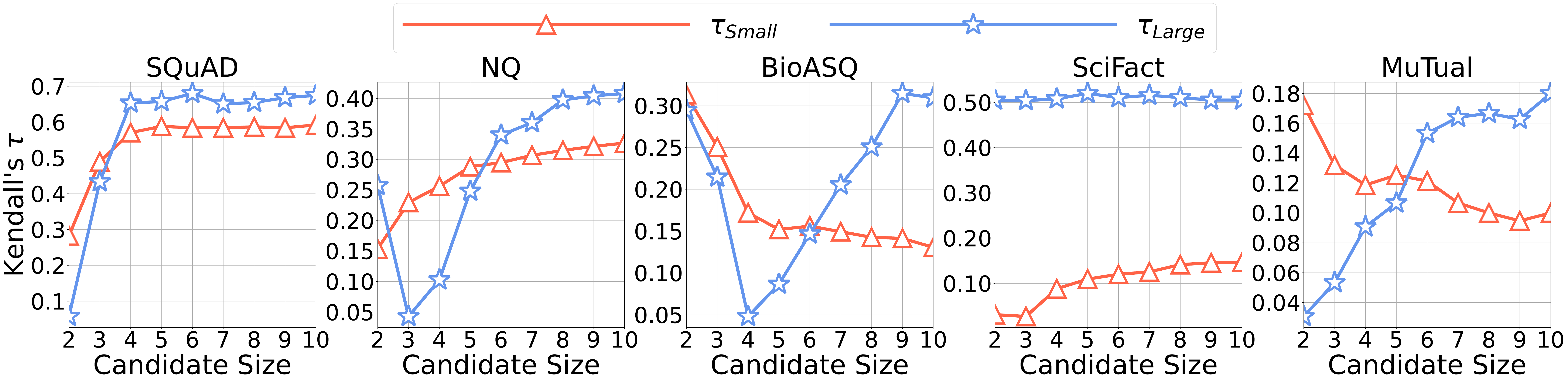}}
    \subfigure[$\mathcal{N}$LEEP]{\label{fig: variation_nleep}\includegraphics[width=2\columnwidth]{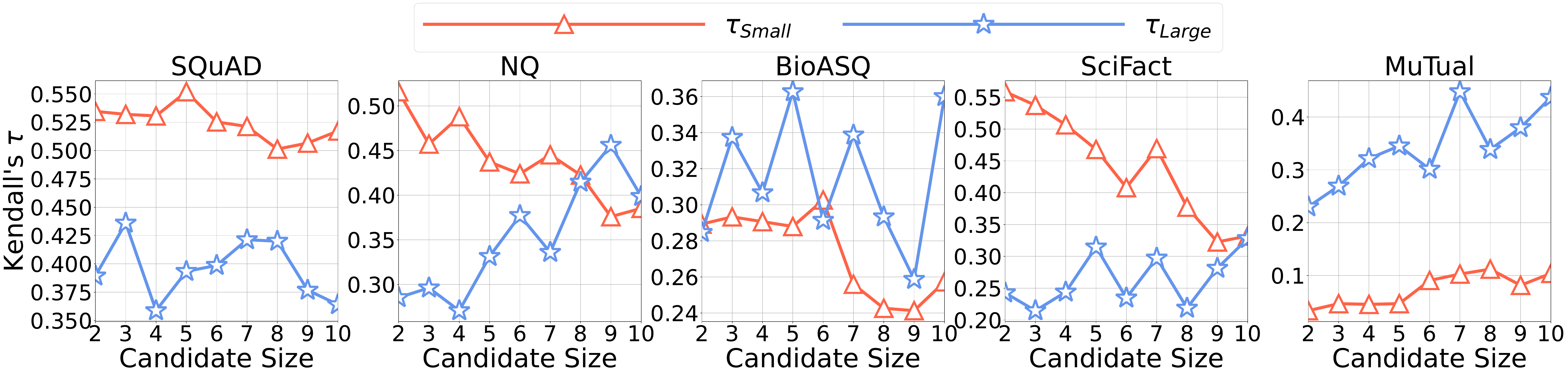}}
    \subfigure[LinearProxy]{\label{fig: variation_linear}\includegraphics[width=2\columnwidth]{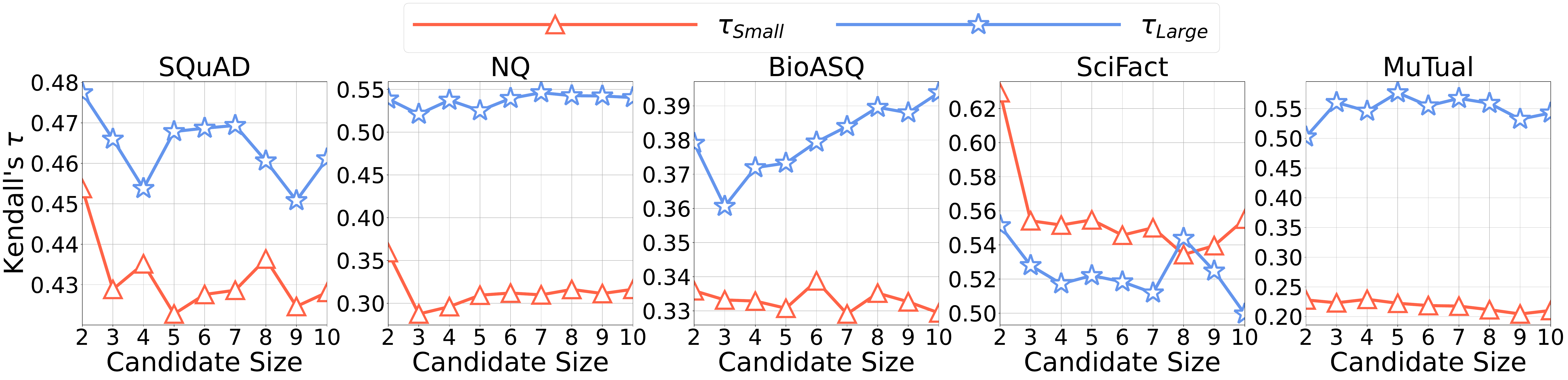}}

  \caption{The performance variations of TMI, GBC, TransRate, $\mathcal{N}$LEEP, and LinearProxy over different sizes of candidate documents on SQuAD, NQ, BioASQ, SciFact, and MuTual datasets.}
  \label{fig: performance_variation_1}
\end{figure*}

\begin{figure*}[t]
  \centering
    \subfigure[SFDA]{\label{fig: variation_sfda}\includegraphics[width=2\columnwidth]{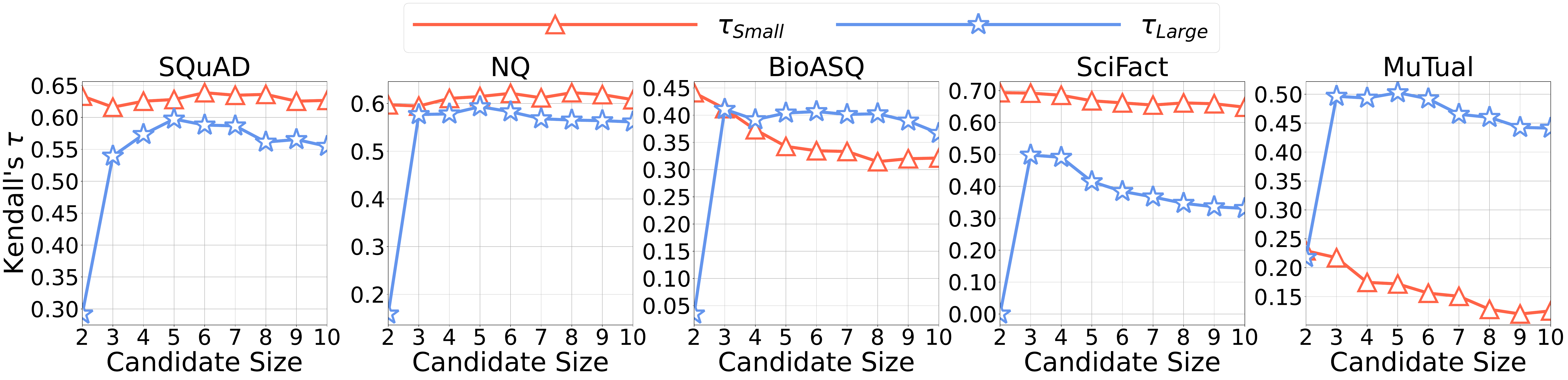}}
    \subfigure[PACTran]{\label{fig: variation_pactran}\includegraphics[width=2\columnwidth]{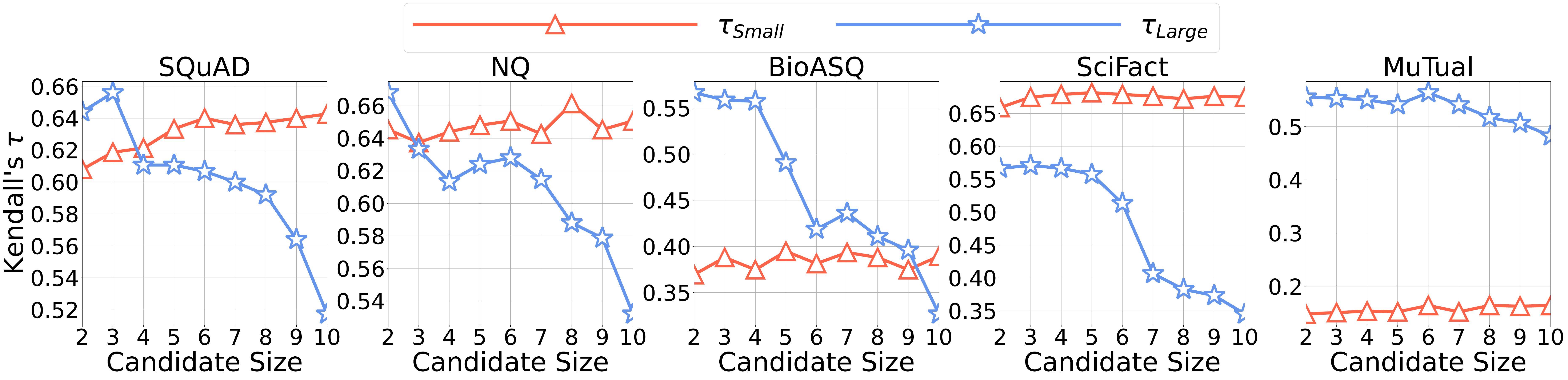}}
    \subfigure[H-score]{\label{fig: variation_hscore}\includegraphics[width=2\columnwidth]{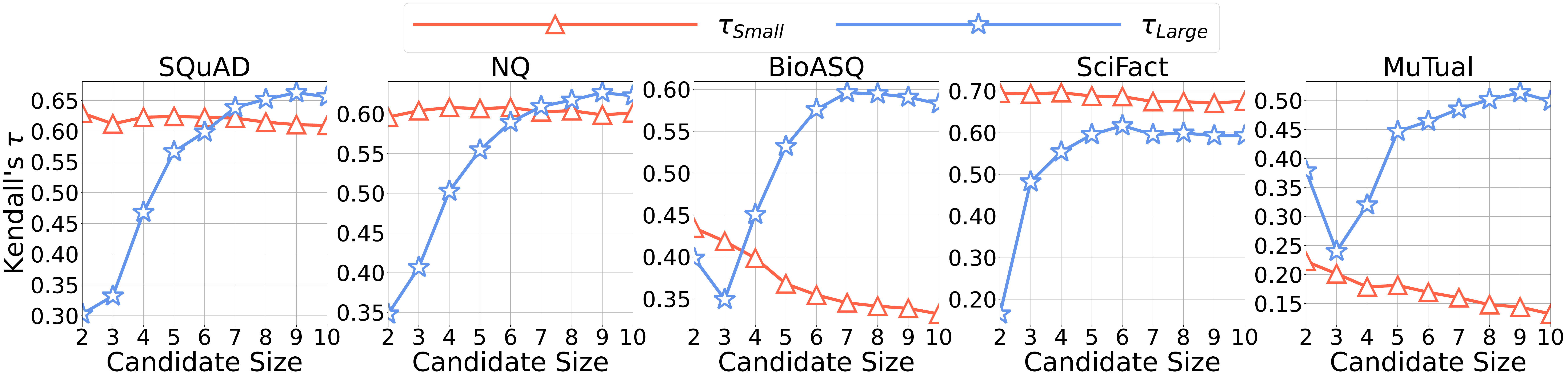}}
    \subfigure[LogME]{\label{fig: variation_logme}\includegraphics[width=2\columnwidth]{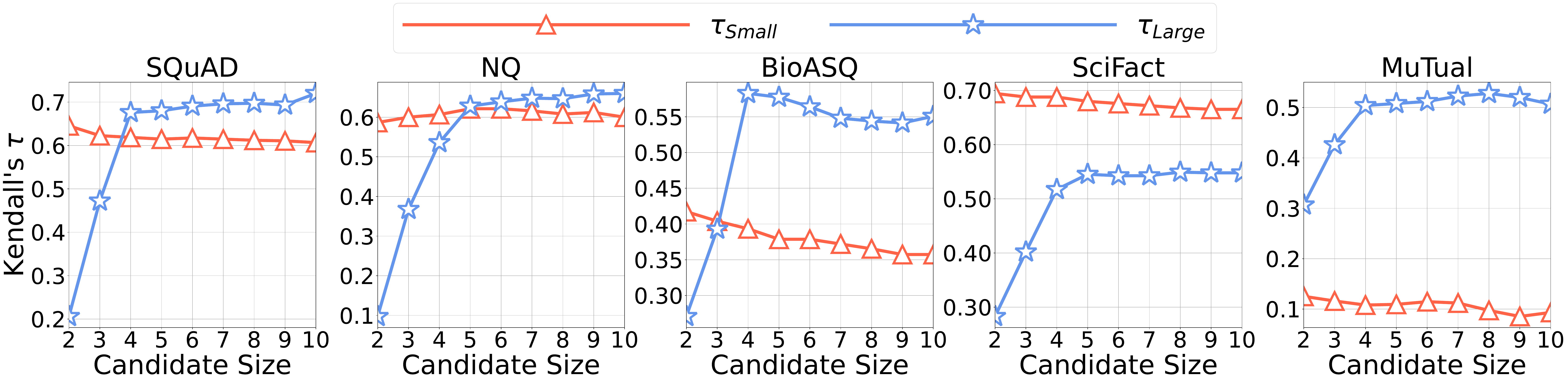}}
    \subfigure[AiRTran]{\label{fig: variation_airtran_}\includegraphics[width=2\columnwidth]{Figs/AiRTran.pdf}}
  
  \caption{The performance variations of SFDA, PACTran, H-score, LogME, and AiRTran over different sizes of candidate documents on SQuAD, NQ, BioASQ, SciFact, and MuTual datasets.}
  \label{fig: performance_variation_2}
\end{figure*}

The performance variations of all TE methods across varying candidate sizes are as depicted in Figures \ref{fig: performance_variation_1} and \ref{fig: performance_variation_2}.

\section{Selection by Human and ChatGPT}
\label{app: chatgpt}
The instruction to guide human experts and ChatGPT to conduct model selection is illustrated in Table \ref{table: instruction_case}. 
Table \ref{table: dataset_meta} and Table \ref{table: model_meta} present the examples of dataset meta-information and model meta-information listed in the instruction, respectively. 
For ChatGPT, all the experiments are conducted 5 times using ChatGPT-4o.

\begin{table*}[t]
    \centering
    \resizebox{0.9\textwidth}{!}{
    \setlength{\tabcolsep}{9.5pt}{
    \begin{tabular}{lccccc}
        \toprule
        \textbf{PLMs} & \textbf{SQuAD} & \textbf{NQ} & \textbf{BioASQ} & \textbf{SciFact} & \textbf{MuTual} \\
        \midrule
        \multicolumn{6}{c}{Small PLMs (Full Fine-tuning)} \\
\hline
        bert-base-uncased & 0.888 & 0.766 & 0.588 & 0.515 & 0.262 \\
bert-base-cased & 0.874 & 0.733 & 0.572 & 0.505 & 0.268 \\
roberta-base & 0.884 & 0.751 & 0.508 & 0.476 & 0.277 \\
biobert-base-cased-v1.1 & 0.851 & 0.704 & 0.633 & 0.565 & 0.244 \\
electra-base-discriminator & 0.807 & 0.717 & 0.525 & 0.285 & 0.212 \\
unsup-simcse-bert-base-uncased & 0.888 & 0.764 & 0.591 & 0.498 & 0.252 \\
sup-simcse-bert-base-uncased & 0.887 & 0.766 & 0.603 & 0.513 & 0.275 \\
openai-gpt & 0.848 & 0.714 & 0.483 & 0.480 & 0.195 \\
bart-base & 0.899 & 0.656 & 0.563 & 0.521 & 0.285 \\
scibert\_scivocab\_cased & 0.850 & 0.703 & 0.617 & 0.527 & 0.222 \\
scibert\_scivocab\_uncased & 0.856 & 0.727 & 0.614 & 0.581 & 0.232 \\
distilbert-base-cased & 0.863 & 0.727 & 0.567 & 0.459 & 0.249 \\
distilbert-base-uncased & 0.883 & 0.757 & 0.589 & 0.512 & 0.255 \\
ernie-2.0-base-en & 0.897 & 0.772 & 0.590 & 0.547 & 0.284 \\
distilroberta-base & 0.882 & 0.737 & 0.517 & 0.449 & 0.268 \\
distilgpt2 & 0.818 & 0.686 & 0.350 & 0.047 & 0.093 \\
distilbert-base-multilingual-cased & 0.858 & 0.720 & 0.513 & 0.462 & 0.225 \\
albert-base-v2 & 0.837 & 0.708 & 0.516 & 0.385 & 0.188 \\
PubMedBERT-base-uncased & 0.856 & 0.723 & 0.633 & 0.609 & 0.245 \\
BioLinkBERT-base & 0.853 & 0.724 & 0.600 & 0.619 & 0.243 \\
sentence-t5-base & 0.887 & 0.754 & 0.461 & 0.293 & 0.229 \\
gte-base & 0.920 & 0.791 & 0.574 & 0.650 & 0.295 \\
contriever & 0.916 & 0.782 & 0.524 & 0.596 & 0.266 \\
e5-base & 0.923 & 0.792 & 0.571 & 0.641 & 0.288 \\
bge-base-en & 0.919 & 0.793 & 0.587 & 0.654 & 0.297 \\
\hline
\multicolumn{6}{c}{Large PLMs (Parameter-efficient Fine-tuning)} \\
\hline
falcon-rw-1b & 0.763 & 0.632 & 0.380 & 0.234 & 0.089 \\
SGPT-1.3B-mean-nli & 0.764 & 0.622 & 0.378 & 0.279 & 0.149 \\
cosmo-1b & 0.737 & 0.565 & 0.329 & 0.126 & 0.065 \\
gpt-neo-1.3B & 0.720 & 0.543 & 0.367 & 0.194 & 0.078 \\
TinyDolphin-2.8-1.1b & 0.764 & 0.643 & 0.368 & 0.179 & 0.105 \\
opt-1.3b & 0.786 & 0.649 & 0.369 & 0.201 & 0.099 \\
Qwen1.5-1.8B & 0.750 & 0.583 & 0.378 & 0.200 & 0.078 \\
bloom-1b7 & 0.635 & 0.128 & 0.267 & 0.010 & 0.036 \\
deepseek-coder-1.3b-base & 0.598 & 0.362 & 0.140 & 0.021 & 0.037 \\
pythia-1.4b & 0.764 & 0.596 & 0.418 & 0.291 & 0.090 \\
phi-1\_5 & 0.774 & 0.592 & 0.388 & 0.178 & 0.095 \\
TinyLlama-1.1B & 0.773 & 0.651 & 0.370 & 0.183 & 0.097 \\
Yuan2-2B-hf & 0.703 & 0.483 & 0.352 & 0.091 & 0.074 \\
stable-code-3b & 0.696 & 0.496 & 0.344 & 0.188 & 0.069 \\
Apollo-2B & 0.590 & 0.494 & 0.194 & 0.121 & 0.071 \\
gemma-2b & 0.527 & 0.447 & 0.236 & 0.099 & 0.043 \\
phi-2 & 0.796 & 0.638 & 0.428 & 0.240 & 0.091 \\
EMO-2B & 0.641 & 0.508 & 0.323 & 0.186 & 0.059 \\
rocket-3B & 0.708 & 0.620 & 0.284 & 0.249 & 0.105 \\
stablelm-3b-4e1t & 0.752 & 0.642 & 0.431 & 0.321 & 0.086 \\
gemma-7b & 0.405 & 0.278 & 0.160 & 0.150 & 0.019 \\
chatglm3-6b & 0.762 & 0.566 & 0.417 & 0.370 & 0.093 \\
Meta-Llama-3-8B & 0.800 & 0.672 & 0.420 & 0.371 & 0.090 \\
Qwen1.5-7B & 0.757 & 0.645 & 0.412 & 0.298 & 0.089 \\
Mistral-7B-v0.3 & 0.783 & 0.649 & 0.392 & 0.357 & 0.082 \\
        \bottomrule
        
    \end{tabular}
    }
    }
    \caption{The best fine-tuning performance of selected PLMs.}
    \label{tab: fine_tuning_results}
\end{table*}

\begin{table*}[h]
	\begin{center}
        \setlength{\tabcolsep}{5pt}{
			\begin{tabular}{m{\textwidth}}
				\hline
                \multicolumn{1}{c}{\textbf{----- Instruction for Model Selection -----}} \\
                \hline
                \textbf{Background}:\\
                The number of open-access Pre-trained Language Models (PLMs) is growing fast in recent years. When a dataset is given, one usually doesn't know which PLM should be selected to achieve the best fine-tuning performance. Since fine-tuning all PLMs to solve this problem is not practical, there is a need of efficient model selection. \\
\textbf{Instructions}:\\
You will receive the detailed meta-information for the specified dataset. Following that, you will receive the meta-information for each candidate model in the model pool. Your task is to score and rank these models based on their suitability for the given dataset, using the provided meta-information as the basis for your decisions, where the score is between 0 and 1, the higher the score, the better the fine-tuning performance of the model. After ranking the models, you must provide a detailed explanation for your rankings, citing specific aspects of the meta-information that influenced your decisions.\\
\textbf{The dataset meta-information is}:\\
...\\
\textbf{The model meta-information is}:\\
(1) ...\\
(2) ...\\
...\\
\textbf{Output Format Requirement is}:\\
You must output ranking results in the form of JSON to represent the order of the candidate models. The JSON format should be:\\
\{\\
  \quad\quad"dataset\_info": \{\\
    \quad\quad\quad\quad"dataset\_name": "Dataset Name",\\
    \quad\quad\quad\quad"description": "Brief description of the dataset."\\
  \quad\quad\},\\
  \quad\quad"model\_ranking": [\\
    \quad\quad\quad\quad\{\\
      \quad\quad\quad\quad\quad\quad"rank": 1,\\
      \quad\quad\quad\quad\quad\quad"model": "Model Name",\\
      \quad\quad\quad\quad\quad\quad"score": "Score Value",\\
      \quad\quad\quad\quad\quad\quad"explanation": "Detailed explanation for the ranking."\\
    \quad\quad\quad\quad\},\\
    \quad\quad\quad\quad\{\\
      \quad\quad\quad\quad\quad\quad"rank": 2,\\
      \quad\quad\quad\quad\quad\quad"model": "Model Name",\\
      \quad\quad\quad\quad\quad\quad"score": "Score Value",\\
      \quad\quad\quad\quad\quad\quad"explanation": "Detailed explanation for the ranking."\\
    \quad\quad\quad\quad\},\\
    \quad\quad\quad\quad\{\\
      \quad\quad\quad\quad\quad\quad"rank": 3,\\
      \quad\quad\quad\quad\quad\quad"model": "Model Name",\\
      \quad\quad\quad\quad\quad\quad"score": "Score Value",\\
      \quad\quad\quad\quad\quad\quad"explanation": "Detailed explanation for the ranking."\\
    \quad\quad\quad\quad\},\\
    \quad\quad\quad\quad...\\
  \quad\quad]\\
\}\\
				\hline
			\end{tabular}}
	\end{center}
    \caption{Instruction for model selection using ChatGPT.}
	\label{table: instruction_case}
\end{table*}

\begin{table*}[h]\
	\begin{center}\
        \setlength{\tabcolsep}{5pt}{
			\begin{tabular}{m{0.9\textwidth}}
				\hline
                \multicolumn{1}{c}{\textbf{----- Meta-information of BioASQ Dataset -----}} \\
                \hline
  \textbf{Dataset Name}: BioASQ. \\
  \hline
  \textbf{Description}: BioASQ is designed to facilitate research in biomedical answer retrieval by transforming existing BioASQ datasets into a format suitable for ReQA tasks...\\
  \hline
  \textbf{Task}: Given a question, the target is to retrieve corresponding answer from answer candidates.\\
  \hline
  \textbf{Source}: Question-answer pairs annotated on PubMed by biomedical experts.\\
  \hline
  \textbf{Data Size}: 3742 questions and 5828 answers in train set, 496 questions and 31682 candidates in test set.\\
  \hline
  \textbf{Language}: English.\\
  \hline
  \textbf{Metric}: P@1, Precision at 1, a measure of how often the top-ranked result is correct.\\
  \hline
  \textbf{Benchmark Performance}: The P@1 of Dual-BioBERT is 57.92, ...\\
  \hline
  \textbf{Related Papers}:\\ 
  $[1]$ Jun Bai, Chuantao Yin, Zimeng Wu, Jianfei Zhang, Yanmeng Wang, Guanyi Jia, Wenge Rong, and Zhang Xiong. 'Improving Biomedical ReQA With Consistent NLI-Transfer and Post-Whitening.' IEEE/ACM Transactions on Computational Biology and Bioinformatics, vol. 20, no. 3, 2023.\\
  \hline
  \textbf{Data Sample Cases}: \\
   "Question 1": "Are gut microbiota profiles altered by irradiation?",\\
   "Answer 1": "Yes, Irradiation profoundly impacted gut microbiota profiles"\\
    ...\\
				\hline
			\end{tabular}}
	\end{center}
    \caption{The example of dataset's meta-information.}
	\label{table: dataset_meta}
\end{table*}

\begin{table*}[h]\
	\begin{center}\
        \setlength{\tabcolsep}{5pt}{
			\begin{tabular}{m{0.9\textwidth}}
				\hline
                \multicolumn{1}{c}{\textbf{----- Meta-information of LLaMA-3-8B -----}} \\
                \hline
  \textbf{Model Name}: meta-llama/Meta-Llama-3-8B.\\
  \hline
  \textbf{Model Structure}:    \\
  \hline   
   Auto-regressive language model, decoder-only structure, 32 layers, grouped-query attention, context length is 8k, hidden size is 4096, totally 8B parameters.\\
   \hline
   \textbf{Pre-training Details}:\\
    Pre-training by causal language modeling on Over 15T tokens from publicly available sources, including a mix of high-quality non-English data to cover over 30 languages.\\
    \hline
    \textbf{Model Performance}:\\
     The Meta-Llama 3-8B model demonstrates impressive performance across various benchmarks, significantly surpassing several competing models. On the MMLU 5-shot benchmark, Meta-Llama 3-8B scored 66.6, outperforming Mistral 7B by 3.1 points and Gemma 7B by 2.3 points, and achieving a remarkable 20.9-point improvement over Llama 2-7B. In the AGIEval English 3-5 shot test, it achieved 45.9, which is 4.2 points higher than Gemma 7B and notably surpasses Llama 2-7B's performance. On the BIG-Bench Hard 3-shot, CoT benchmark, Meta-Llama 3-8B scored 61.1, showing a substantial improvement of 23.1 points over Llama 2-13B and outperforming Mistral 7B by 5.1 points. In the ARC-Challenge 25-shot test, Meta-Llama 3-8B achieved a score of 78.6, outperforming Gemma 7B by 25.4 points and Llama 2-13B by 11.0 points. Lastly, on the DROP 3-shot, F1 benchmark, Meta-Llama 3-8B scored 58.4, which is 4.0 points higher than Gemma 7B and significantly outperforms Llama 2-7B by 20.5 points. These results underscore Meta-Llama 3-8B's advanced capabilities and superior performance in natural language processing tasks compared to its predecessors and similar models.\\
				\hline
			\end{tabular}}
	\end{center}
    \caption{The example of model's meta-information.}
	\label{table: model_meta}
\end{table*}

\end{document}